\documentclass[pmlr]{jmlr}

\usepackage{longtable}
\usepackage{pgfplots}

 \usepackage{booktabs}
 
\usepackage[load-configurations=version-1]{siunitx} 
\usepackage{tabulary}
\makeatletter
\def\set@curr@file#1{\def\@curr@file{#1}} 
\makeatother

\usepackage{cleveref}

\newcommand{\eat}[1]{\ignorespaces}

\theorembodyfont{\upshape}
\theoremheaderfont{\scshape}
\theorempostheader{:}
\theoremsep{\newline}

\jmlrvolume{}
\jmlryear{2023}
\jmlrworkshop{}

\usepackage[nolist]{acronym}

\makeatletter
\newcommand*{\org@overidelabel}{}
\let\org@overridelabel\@verridelabel
\@ifpackagelater{acronym}{2015/03/21}{
  \renewcommand*{\@verridelabel}[1]{%
    \@bsphack
    \protected@write\@auxout{}{\string\AC@undonewlabel{#1@cref}}%
    \org@overridelabel{#1}%
    \@esphack
  }%
}{
  \renewcommand*{\@verridelabel}[1]{%
    \@bsphack
    \protected@write\@auxout{}{\string\undonewlabel{#1@cref}}%
    \org@overridelabel{#1}%
    \@esphack
  }%
}
\makeatother

\title[Distilling LLMs for Biomedical Knowledge Extraction]{Distilling Large Language Models for Biomedical Knowledge Extraction: A Case Study on Adverse Drug Events}

\author{\Name{Yu Gu}, \Name{Sheng Zhang}, \Name{Naoto Usuyama}, \Name{Yonas Woldesenbet}, \Name{Cliff Wong}, \Name{Praneeth Sanapathi}, \Name{Mu Wei}, \Name{Naveen Valluri}, \Name{Erika Strandberg}, \Name{Tristan Naumann}, \Name{Hoifung Poon} \\
\addr Microsoft Research}

\begin{document}

\maketitle

\begin{acronym}
    \acro{ADE}{adverse drug event}
    \acro{AE}{adverse event}
    \acro{LLM}{large language model}
    \acro{NER}{named entity recognition}
    \acro{NLP}{natural language processing}
    \acro{PHI}{protected health information}
    \acro{RE}{relation extraction}
\end{acronym}

\begin{abstract}
  Large language models (LLMs), such as GPT-4, have demonstrated remarkable capabilities across a wide range of tasks, including health applications. In this paper, we study how LLMs can be used to scale biomedical knowledge curation. We find that while LLMs already possess decent competency in structuring biomedical text, by distillation into a task-specific student model through self-supervised learning, substantial gains can be attained over out-of-box LLMs, with additional advantages such as cost, efficiency, and white-box model access.
  We conduct a case study on adverse drug event (ADE) extraction, which is an important area for improving care. On standard ADE extraction evaluation, a GPT-3.5 distilled PubMedBERT model attained comparable accuracy as supervised state-of-the-art models without using any labeled data. Despite being over 1,000 times smaller, the distilled model outperformed its teacher GPT-3.5 by over 6 absolute points in F1 and GPT-4 by over 5 absolute points.
  Ablation studies on distillation model choice (e.g., PubMedBERT vs BioGPT) and ADE extraction architecture shed light on best practice for biomedical knowledge extraction. Similar gains were attained by distillation for other standard biomedical knowledge extraction tasks such as gene-disease associations and protected health information, further illustrating the promise of this approach.
\end{abstract}
\section{Introduction}

\Acp{ADE} pose a significant public health challenge because they represent injuries resulting from medical interventions related to drug use, including medication errors, adverse drug reactions, allergic reactions, and overdoses~\citep{donaldson2000err}. In the United States, \acp{ADE} are prevalent and are considered to be among the leading causes of increased mortality, extended hospital stays, and elevated healthcare costs~\citep{classen1997adverse}. Curating \acp{ADE} from biomedical text is thus essential to ensuring and improving patient safety, but remains expensive and time consuming because it is predominantly done manually.~\citep{chen2020extracting}.

Automated systems for evidence-based pharmacovigilance can help address the challenges of manual \ac{ADE} identification, particularly for pharmaceutical and healthcare companies~\citep{gurulingappa2012development}. However, constructing a gold standard corpus for \ac{ADE} identification remains challenging due to the need for multiple specialized annotators with extensive biomedical backgrounds.

\Acp{LLM}, such as GPT-4, have demonstrated impressive zero-shot and few-shot capabilities in both general domains~\citep{openai2023gpt4,bubeck2023sparks} and health applications~\citep{lee2023ai}. 
In this paper, we study how \acp{LLM} can be leveraged to scale biomedical knowledge extraction, using \acp{ADE} curation as a case study. 
Our study revealed that state-of-the-art \acp{LLM}, such as GPT-3.5 or GPT-4, already perform competitively in ADE extraction in zero-shot or few-shot settings, but still trail state-of-the-art supervised systems by a large margin.
Interestingly, by leveraging \acp{LLM} as a noisy teacher to annotate large unlabeled data, we can distill its  capabilities into a task-specific student model that is not only more efficient, but also substantially outperforms the teacher model in end applications.
On standard \ac{ADE} extraction evaluation, PubMedBERT~\citep{gu2021domain} distilled from GPT-3.5 attained comparable accuracy as supervised state-of-the-art models without using any labeled examples. Despite being over 1,000 times smaller, the distilled model outperformed its noisy teacher GPT-3.5 by over six~(6) absolute points in F1 and GPT-4 by over five~(5) absolute points.
Unlike GPT-3.5 or GPT-4, such a distilled model offers white-box access and can be further fine-tuned or customized for specialized uses.

We found similar gains from \ac{LLM} distillation for other standard biomedical knowledge extraction tasks such as gene-disease associations and \ac{PHI}, further illustrating the promise of this approach. 
We also conduct ablation studies on key distillation design such as neural architecture and model choice, which help establish best practice for biomedical knowledge extraction. To facilitate future research in this direction, we will release our distilled models.

\subsection*{Generalizable Insights about Machine Learning in the Context of Healthcare}
\begin{itemize}
\item Knowledge distillation from \acp{LLM} and self-supervision techniques boost the performance of information extraction tasks in the biomedical domain, which provides a general and reliable solution to various healthcare applications.

\item The proposed end-to-end architecture for \ac{ADE} extraction underscores the importance of adapting machine learning models to the unique challenges and requirements of healthcare-related problems, increasing their relevance and impact in clinical settings.

\item The successful application of our approach to \ac{ADE} extraction emphasizes the potential for transferring knowledge from \acp{LLM} to other natural language processing tasks in healthcare, contributing to a broader understanding of machine learning techniques in this domain.

\end{itemize}

\section{Related Work}
There are two key areas of related work: end-to-end \ac{ADE} extraction and knowledge distillation.

\subsection{End-to-end ADE Extraction}
A variety of approaches have been proposed for \ac{ADE} extraction. Among these, SpERT~\citep{eberts2019span} utilizes lightweight reasoning on BERT embeddings for joint entity and relation extraction, demonstrating the potential for combining these tasks. REBEL~\citep{cabot2021rebel}, an autoregressive seq2seq model based on BART, simplifies relation extraction by representing triplets as text sequences and achieves state-of-the-art performance on multiple benchmarks. The table-sequence encoder model \citep{wang2020two} employs two distinct encoders to capture different information types during the learning process, showcasing significant improvements over existing single-encoder approaches.

\subsection{Knowledge Distillation}
Earlier \acp{LLM}, such as GPT-3 \citep{ouyang2022training, agrawal2022large}, demonstrated great potential but fell short of competitive results on biomedical \ac{NLP} tasks \citep{gutierrez2022thinking, moradi2022gpt3}. However, the creation of GPT-3.5 and GPT-4 \citep{openai2023gpt4}, the latest generation of domain-agnostic \acp{LLM}, has generated new opportunities for advancing medicine, health, and public understanding of the capabilities and limitations of these models \citep{lee2023ai}.

In this work, we concentrate on knowledge distillation of LLMs using self-supervision techniques \citep{agrawal2022large, smith2022language}. In other words, we use these \acp{LLM} as labelers in the biomedical domain, capitalizing on their powerful language understanding capabilities to generate high-quality labels for various tasks. Our experiments highlight the advantages of this approach for enhancing performance on challenging biomedical \ac{NLP} tasks, especially \ac{ADE} extraction, illustrating the potential of self-supervised distillation for harnessing the power of state-of-the-art \acp{LLM} in specialized domains.

\section{Methods}



\subsection{Task Definition}

In this study, we focus on end-to-end \ac{ADE} extraction, which involves two separate \ac{NLP} sub-tasks: (1) identifying \ac{AE} mentions using \ac{NER}, where a drug causation is not yet assigned, and (2) assigning causation to drugs through \ac{RE}, which aims to find the relations between \acp{AE} and corresponding drugs.

The first sub-task, \ac{AE} entity extraction, focuses on locating and identifying mentions of adverse events within the given text. This step is crucial for gathering information about potential negative effects associated with drugs, without considering causation at this stage.

The second sub-task, \ac{ADE} relation extraction, aims to establish causal links between the extracted \ac{AE} entities and drugs in the context. This step is essential for understanding the relationships between drugs and their adverse effects, enabling more informed decisions regarding drug safety and usage.

To validate our proposed method, we utilize the \ac{ADE} corpus \citep{gurulingappa2012development}, a dataset systematically annotated for supporting the automatic extraction of drug-related adverse effects from medical reports. This dataset allows us to evaluate the performance of our approach on both subtasks, providing a comprehensive assessment of the end-to-end \ac{ADE} extraction process.


\subsection{A Unified Neural Architecture for ADE Extraction}

\begin{figure}[t]
\centering
\includegraphics[width=\textwidth]{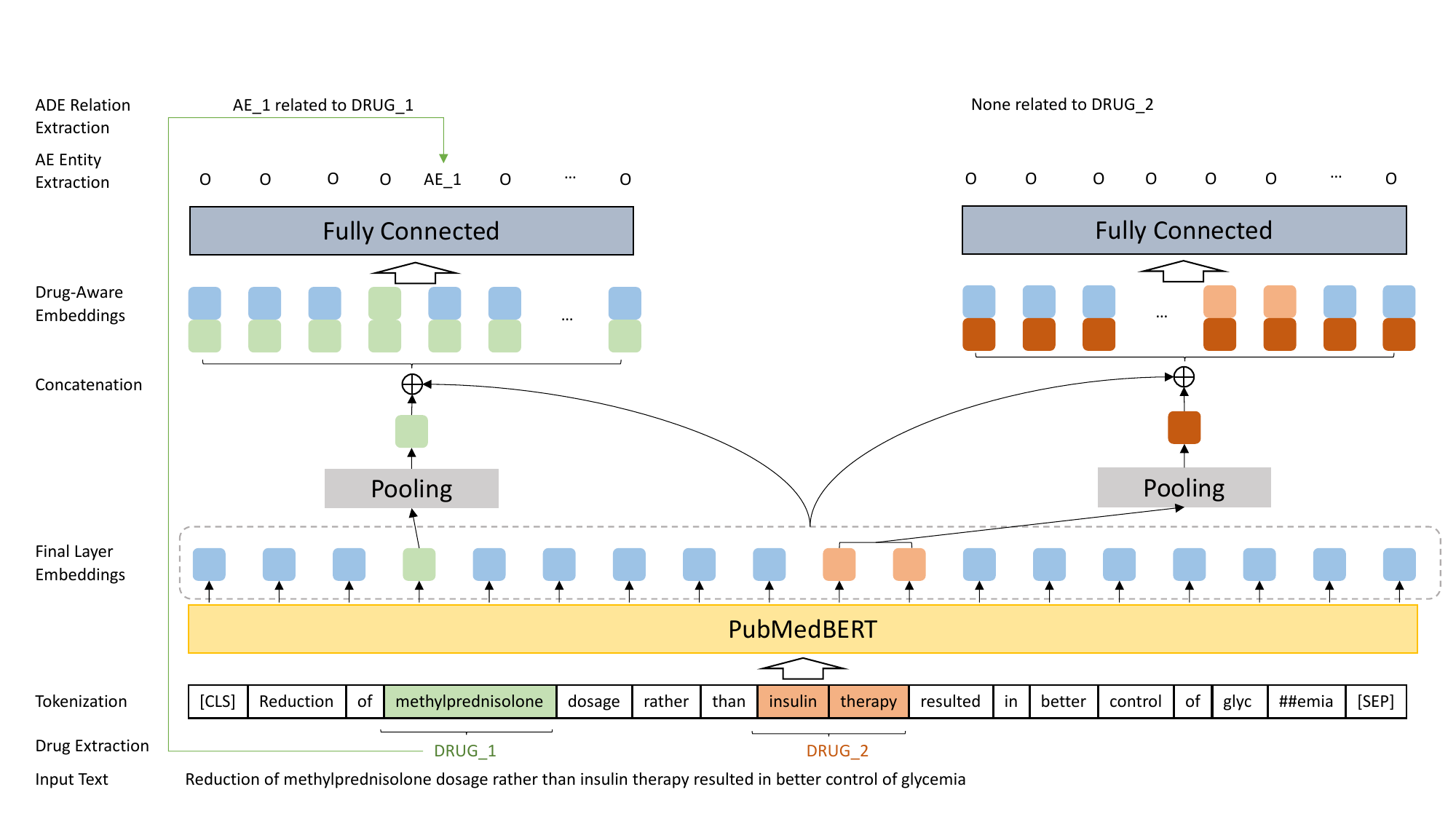}
\caption{Our unified neural architecture for extracting adverse-event arguments and assigning causation relations for each drug entity in question (DRUG\_1 and DRUG\_2 in this case). The input sequence is first passed through an encoder (PubMedBERT) and then augmented by concatenation with the drug representation, which is generated by mean-pooling the encoding of all mention tokens. A linear fully connected layer is then applied for token classification using softmax, predicting adverse event tokens pertaining to the designated drug entity. This architecture significantly reduces computational complexity from enumerating all pairwise combinations of adverse events and drugs to only enumerating drug entities, facilitating efficient and accurate adverse drug event extraction.}
\label{fig:unified_arch}
\end{figure}


Traditional methods for \ac{ADE} extraction typically treat the two subtasks, AE identification (NER) and ADE relation extraction (RE), as separate processes. However, in situations where multiple \acp{AE} ($N$ mentions) and drugs ($M$ mentions) coexist in the same context, this approach necessitates $\mathcal{O}(NM)$ inferences, leading to a bottleneck for large-scale processing.

Recent studies attempt to tackle this challenge by jointly extracting drug and ADE entities, even though \emph{drug extraction} has been largely addressed in prior work\citep{santosh2021joint, cabot2021rebel}. In this paper, we propose a novel unified architecture that concentrates on efficient and precise extraction of ADE entities and causation assignment. Our model introduces a drug-centric structure, designed to simultaneously handle ADE NER and relation extraction in one pass.

As illustrated in \autoref{fig:unified_arch}, the input sequence undergoes processing to obtain the final layer hidden state output for each drug entity. Denote the input sequence as $x = {x_1, x_2, \ldots, x_T}$, where $x_i$ is the $i$-th token, and $T$ is the sequence length. The output of the final layer hidden state is represented as $H = {h_1, h_2, \ldots, h_T}$, where $h_i \in \mathbb{R}^d$ is the $d$-dimensional hidden state corresponding to the $i$-th token.

We then create a new input sequence for each drug entity. Given a set of drug entities $D = {d_1, d_2, \ldots, d_M}$, where $d_j$ is the $j$-th drug entity, for each drug, hidden states of drug entity are mean-pooled. The resulting pooled token $\bar{d}_j$ is concatenated to every hidden state output token of the input sequence, effectively integrating drug information into each token:

\begin{equation}
\tilde{h}_{j,i} = \text{concat}(h_i, \bar{d}_j)
\end{equation}

where $\tilde{h}_{j,i} \in \mathbb{R}^{2d}$ is the concatenated hidden state for the $i$-th token in the new input sequence created for the $j$-th drug entity.

Subsequently, a linear layer is applied on top of the concatenated tokens for binary token classification using sigmoid. This process transforms the task into predicting ADE tokens while considering the causation drugs. The linear layer and sigmoid are defined as:

\begin{equation}
z_{j,i} = W \tilde{h}_{j,i} + b
\end{equation}

\begin{equation}
p_{j,i} = \sigma(z_{j,i}) = \frac{1}{1 + \exp(-z_{j,i})}
\end{equation}

where $W \in \mathbb{R}^{d'}$ and $b \in \mathbb{R}$ are learnable parameters of the linear layer, with $d' = 2d$ being the dimensionality of the concatenated hidden states, and $p_{j,i}$ represents the predicted probability of the $i$-th token in the new input sequence created for the $j$-th drug entity being an ADE mention.

The proposed architecture substantially simplifies the problem, converting the original two tasks (NER and RE) into a single, unified task. As a result, the computational requirement is dramatically reduced from $\mathcal{O}(NM)$ (all pairwise combinations of adverse events and drugs) to $\mathcal{O}(M)$ (all drug entities), enabling our end-to-end model to perform more efficiently and accurately in large-scale \ac{ADE} extraction.

\subsection{Knowledge Distillation from LLMs}

\begin{figure}[t]
\centering
\includegraphics[width=0.8\textwidth]{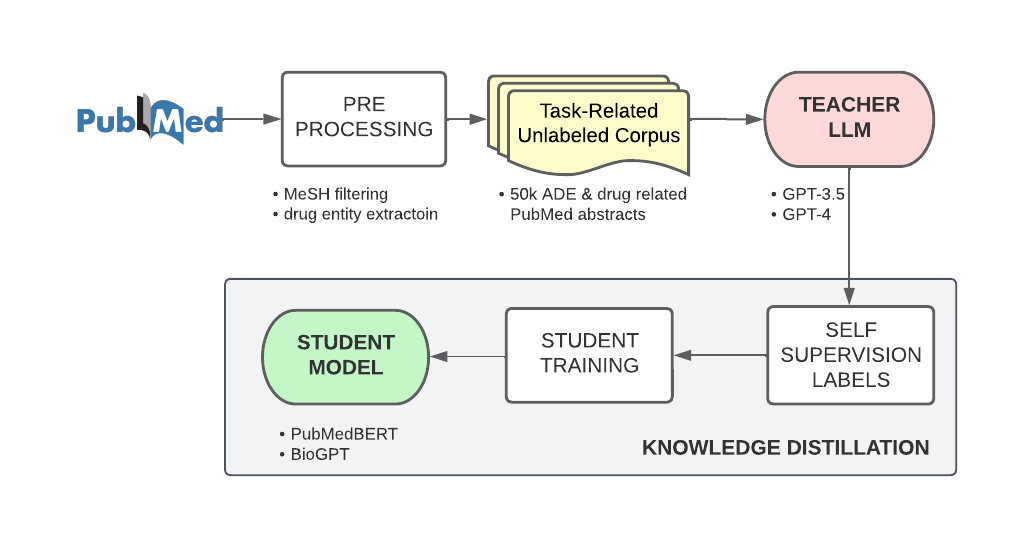}
\caption{Our knowledge distillation framework for self-supervising ADE extraction using LLMs. We first filter PubMed abstracts and select drug-related ones as the unlabeled corpus for ADE self-supervision. We then call upon the teacher LLM (e.g., GPT-3.5) to generate ADE annotations and train a student model.}
\label{fig:distillation}
\end{figure}


We employ knowledge distillation (see Figure \ref{fig:distillation}) using GPT-3.5 as the teacher model.

\subsubsection{Data Curation and Preprocessing}

We adapt the methodology from \cite{gurulingappa2012development} to curate a corpus focused on drug-related adverse events. First, we perform a PubMed search with ``drug therapy" and ``adverse effects" as MeSH terms, limiting the language to English. This search yields approximately 50,000 PubMed abstracts related to drug-related adverse events. The query is as follows:
\begin{quote}
``adverse effects"[sh] AND (hasabstract[text] AND Case Reports[ptyp]) AND ``drug therapy"[sh] AND English[lang] AND (Case Reports[ptyp])
\end{quote}

To efficiently recognize and normalize drug names in the abstracts, we compile a list of oncology drug names, synonyms, and abbreviations from the NCI Thesaurus. We construct a trie from this list for rapid search and identification within plain text. Next, we split each abstract into sentences, retaining only those containing identified drug names. This process results in a refined ADE related dataset suitable for knowledge distillation.

\subsubsection{Teacher and Student Models in Knowledge Distillation}

Our knowledge distillation process involves two models: the teacher model, which serves as the source of self-supervision, and the student model, which learns from self-supervised labels produced by the teacher model.

\noindent\textbf{Teacher LLM}\
We employ GPT-3.5 \citep{ouyang2022training} as our teacher model. This advanced language model has demonstrated remarkable performance across various \ac{NLP} tasks, showcasing its strong understanding and reasoning capabilities. To access GPT-3.5, we utilize Azure OpenAI Service, which allows us to interact with the model efficiently and securely. Through the API, we can submit input prompts and receive generated responses, from which we will generate self-supervised data to train our student model.

\noindent\textbf{Student Models} We consider the following state-of-the-art pretrained models for biomedical \ac{NLP}: 1) PubMedBERT \citep{gu2021domain} and PubMedBERT-Large \citep{tinn2021finetuning} are domain-specific language models pretrained on PubMed text; 
2) BioGPT \citep{luo2022biogpt} is a domain-specific generative pretrained transformer model pretrained on PubMed text.

\subsubsection{Knowledge Distillation Process}

We outline the knowledge distillation process, which includes generating input-output pairs, training the student models, and evaluating their performance. 

\noindent\textbf{Generating Input-Output Pairs}
We split our ADE-related unlabeled corpus into sentences and input them to GPT-3.5. We then filter the responses to include only sentences with positive ADE relations, and subsample 40,000 sentences for student model training.

\noindent\textbf{Training the Student Models}
We fine-tune the student models using the generated input-output pairs as labeled examples.
For PubMedBERT, we fine-tune the entire model using our proposed architecture.
For BioGPT, we employ prefix soft tuning~\citep{li-liang-2021-prefix} as standard for GPT models.

\noindent\textbf{Prompt Design}
We experiment with zero-shot and few-shot settings, utilizing in-context learning or prompt-based learning. For the zero-shot setting, we provide a task description in the prompt and instruct the model to return ``none'' if no ADE is found, which helps reduce hallucination. For the few-shot setting, we use the same prompt and add five randomly sampled examples (Figure \ref{fig:ade_prompts}).

\noindent\textbf{Post-Processing}
In pratice, we found that GPT-3.5 and GPT-4 may fail to identifying the exact span of adverse events and often hallucinate non-existing spans. Therefore, we adapt the prompt to ask for the strings only and identify the mentions by string matching.

\noindent\textbf{Evaluation}
We employ the same evaluation metric for both supervised learning and the model-distilled self-supervision approaches, ensuring a fair comparison between the two methods. This metric accounts for the precision, recall, and F1-score, providing a comprehensive assessment of the models' performance in the ADE extraction task.

\begin{figure}[!ht]
\centering
\fbox{%
\parbox{0.9\textwidth}{%
\textbf{Prompt}: Extract the adverse events each drug causes in the Message. If no ADE is found, return None.\

\textit{Example 1:}\\
Message: We postulate that the bolus of sulprostone resulted in possible coronary spasm that resulted in cardiac arrest.\\
Annotations: sulprostone: cardiac arrest$\vert$coronary spasm\\

\textit{Example 2:}\\
Message: In each of the three reported patients, alteration of eyelid appearance with deepening of the lid sulcus was evident as the result of topical bimatoprost therapy.\\
Annotations: bimatoprost: alteration of eyelid appearance$\vert$deepening of the lid sulcus\\

\textit{Example 3:}\\
Message: Immobilization, while Paget's bone disease was present, and perhaps enhanced activation of dihydrotachysterol by rifampicin, could have led to increased calcium - release into the circulation.\\
Annotations: dihydrotachysterol: increased calcium - release\\

\textit{Example 4:}\\
Message: In two patients clozapine was reinstated after risperidone was discontinued; serum triglyceride levels increased.\\
Annotations: clozapine: serum triglyceride levels increased\\

\textit{Example 5:}\\
Message: The cause of these previously unreported side effects of niacin therapy is uncertain but may be related to prostaglandin - mediated vasodilatation, hyperalgesia of sensory nerve receptors, and potentiation of inflammation in the gingiva with referral of pain to the teeth.\\
Annotations: niacin: hyperalgesia of sensory nerve receptors$\vert$pain to the teeth$\vert$potentiation of inflammation in the gingiva$\vert$prostaglandin - mediated vasodilatation
}
}
\caption{Our GPT five-shot prompt for ADE extraction and distillation. The examples are chosen randomly. Our zero-shot prompt is similar, except without the examples.}
\label{fig:ade_prompts}
\end{figure}



\section{Experiments}

\subsection{Evaluation Approach and Study Design} \label{eval_method}

To assess the efficacy of our proposed method, we first provide details on the evaluation approach and study design. The ADE dataset ~\citep{gurulingappa2012development} comprises 6,821 ADE relations in 4,272 sentences. As no official train/dev/test split is provided, we divide the dataset into 8:1:1 for train/dev/test split in our study.

We conduct an end-to-end evaluation wherein the correctness of an ADE is determined only when both entity extraction and its corresponding drug relation are accurate. We report results in terms of lenient F1 score as the primary metric in this study. Lenient F1 score is calculated by considering a true positive when the extracted entity is partially or completely correct, allowing for some flexibility in the boundaries of the extracted entities, while maintaining strict accuracy requirements for the relations between entities. This choice is motivated by the low inter-annotator agreement ratio pertaining to the exact boundaries of ADE entities \citep{henry20202018,gurulingappa2012development}, and our observation of inconsistent mention boundaries of adverse events in the dataset, as detailed in \Cref{sec:annotation-inconsistencies}.

\subsection{ADE Extraction Results}

\Cref{tab:res-distill} compares how various methods perform on ADE extraction: LLM (out-of-box), distillation, supervised. Impressively, out of box, GPT-3.5 and GPT-4 already perform competitively, especially with in-context learning (five-shot). However, they still trail supervised models by a large margin. Interesting, through LLM distillation, a PubMedBERT model already attains comparable accuracy as the supervised state of the art, while using zero labeled example. Although being over three orders of magnitude smaller, this PubMedBERT model outperforms its teacher GPT-3.5 by over six absolute points and outperforms GPT-4 by over five absolute points. Compared with PubMedBERT, the distilled BioGPT performs less well. This is not surprising as it's broadly in line with the observations by \cite{luo2022biogpt}: GPT models are superior for generation tasks such as question answering and summarization, but face more challenges in structuring tasks such as knowledge extraction. We leave more in-depth exploration between GPT and BERT models to future work.

\begin{table}[t]
  \centering 
  \caption{Comparison of LLMs (out-of-box), distillation, and supervised methods on the standard adverse drug event extraction evaluation~\citep{gurulingappa2012development}. Despite of being over 1,000 times smaller, the distilled PubMedBERT model substantially outperforms its teacher LLM (five-shot GPT-3.5) and attains test F1 (lenient) comparable to supervised state of the art. }
 \begin{tabulary}{\textwidth}{l*{2}{>{\raggedright\arraybackslash}J} *{2}{>{\raggedright\arraybackslash}p{2cm}}}
  
  \toprule
  
    \textbf{Method} & \textbf{Teacher LLM} & \textbf{Model} & \textbf{Training Instances}  & \textbf{Test F1}\\
    \midrule
    LLM out-of-box & - & zero-shot GPT-3.5 & - & 78.22 \\ 
    LLM out-of-box & - & zero-shot GPT-4 & - & 84.92 \\ 
    LLM out-of-box & - & 5-shot GPT-3.5 & - & 85.21 \\ 
    LLM out-of-box &- & 5-shot GPT-4 & - & 86.45 \\ 
    \hline
    Distillation & 5-shot GPT-3.5  & BioGPT  & 40,000 & 84.21 \\
    Distillation & 5-shot GPT-3.5 & PubMedBERT  & 40,000 & 91.99 \\
    \hline
    Supervised Learning & - & BioGPT & 3,417 & 88.08 \\ 
    Supervised Learning & - & PubMedBERT & 3,417 & 93.36 \\ 
    \bottomrule
  \end{tabulary}
  \label{tab:res-distill} 

\end{table}

\eat{
\begin{table}[t]
  \centering 
  \caption{Comparison of different methods on the ADE task. GPT-3.5 fewshot is used as teacher model in knowledge distillation process. The proposed task-specific model architecture (see Figure 1) is used in the supervised method. Lenient F1 is used as primary metric. }
  \begin{tabulary}{\textwidth}{l*{5}{>{\raggedright\arraybackslash}J}}
  \toprule
  
    \textbf{Method} & \textbf{Model} & \textbf{Corpus} & \textbf{Training labels}  &\textbf{Training size} & \textbf{Lenient F1}\\
    \midrule
    Supervised & PubMedBERT & ADE corpus & human annotation & 3,417 & 93.36 \\ 
    Distillation & PubMedBERT & PubMed abstracts & GPT-3.5 fewshot & 40,000 & 91.99 \\
    Supervised & BioGPT & ADE corpus & human annotation & 3,417 & 88.08 \\ 
    Few Shot & GPT-4 & ADE corpus & human annotation & 5 & 86.45 \\ 
    Few Shot & GPT-3.5 & ADE corpus & human annotation & 5 & 85.21 \\ 
    Zero Shot & GPT-4 & - & - & 0 & 84.92 \\ 
    Distillation & BioGPT & PubMed abstracts & GPT3.5 fewshot & 40,000 & 84.21 \\
    Zero Shot & GPT-3.5 & - & - & 0 & 78.22 \\ 
    \bottomrule
  \end{tabulary}
  \label{tab:res-distill} 

\end{table}
}

\begin{figure}[t]
\centering
\includegraphics[width=4in]{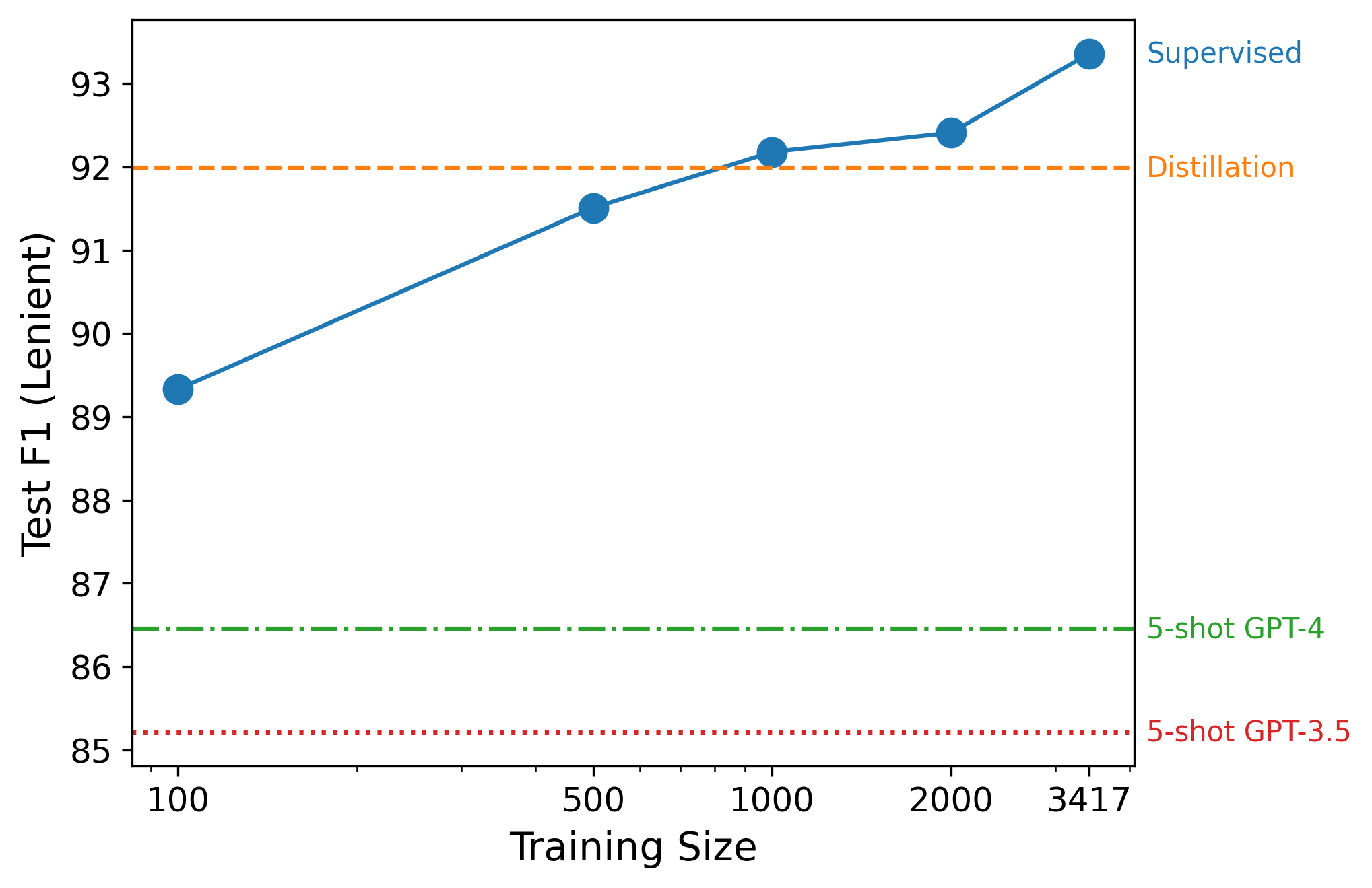}
\caption{Comparison of distillation and low-resource supervised methods on the basis of Lenient F1 scores across various training sizes. As the training size decreases, the performance of the supervised model gradually degrades, with knowledge distillation offering a competitive alternative.}
\label{fig:low_resource_curve}
\end{figure}

Figure \ref{fig:low_resource_curve} shows the supervised learning curve for PubMedBERT on ADE extraction, and how the few-shot LLMs and distillation (also with PubMedBERT) compare. Out of box, LLMs still trail supervised methods by some distance. However, with distillation and without required any labeled data, this gap can be substantially reduced, which bodes well for general applications where we can't afford extensive annotation but still want to attain higher accuracy than the original LLMs. There are also additional benefits, such as cost, efficiency, white-box model access.


\subsection{Comparison on ADE Extraction Models}

\begin{table}[t]
  \centering 
  \caption{Comparison of our proposed neural architecture with prior state-of-the-art methods in the supervised setting on the standard adverse drug event extraction evaluation. To enable head-to-head comparison, we follow prior methods to report strict F1 with 10-fold cross validation. So the numbers are not directly comparable with our other reported results.}
  \begin{tabular}{llllll}
  \toprule
    \textbf{Model} & \textbf{Test F1 (Strict with 10-fold CV)}\\
    \midrule
    SpERT \citep{eberts2019span}& 79.24 \\ 
    Table-Sequence \citep{wang2020two}  & 80.01 \\ 
    SpERT.PL \citep{santosh2021joint} & 82.03 \\ 	
    REBEL \citep{cabot2021rebel} &82.20 \\
    Ours (PubMedBERT)& 84.27 \\ 
    Ours (PubMedBERT-Large)& 84.53 \\ 
    \bottomrule
  \end{tabular}
  \label{tab:comp-prior} 
\end{table}



To compare our propose neural architecture \ref{fig:unified_arch} with prior approaches, we follow prior work to perform 10-fold cross-validation on the ADE corpus and report "strict" F1 scores, where an adverse event entity is deemed correct only when the mention span matches the gold exactly. 
As shown in Table \ref{tab:comp-prior}, our models outperform all prior state of the art, indicating that the proposed neural architecture is advantageous for ADE extraction.


\subsection{LLM Distillation for other Biomedical NLP Tasks}

we evaluate the impact of LLM distillation on other biomedical NLP tasks, as shown in Table \ref{tab:othertasks}. Below is the task description:

\begin{itemize}
\item \textbf{GAD} - The Gene-Disease Association (GAD) \citep{becker2004genetic} task focuses on identifying associations between genes and diseases from biomedical literature. This task requires the extraction of gene and disease entities from text, as well as the determination of their relationships. The performance of models on this task is crucial for understanding genetic influences on diseases and advancing precision medicine.

\item \textbf{PHI (i2b2 2014)} - The Protected Health Information (PHI) task, specifically the i2b2 2014 shared task \citep{uzuner20142014}, aims at identifying and redacting personal identifiers in clinical text. The goal is to remove any information that could be used to trace back to individual patients, ensuring privacy and compliance with regulations such as the Health Insurance Portability and Accountability Act (HIPAA).

\item \textbf{MedNLI} - The Medical Natural Language Inference (MedNLI) \citep{romanov2018lessons} task is based on the NLI task, which involves determining the relationship between a pair of sentences (entailment, contradiction, or neutral). In the context of MedNLI, the sentences are derived from clinical text, making this task valuable for understanding complex relationships in medical documents.

\end{itemize}



\begin{table}[]
 \centering 
  \caption{Comparison of applying GPT-3.5 out-of-box vs. distilling into a PubMedBERT student model on additional biomedical NLP tasks. GAD and PHI are standard biomedical knowledge extraction tasks, whereas MedNLI is a text-entailment task. For simplicity, during distillation, we only use the unlabeled text in the training data of each task (with labels excluded) for LLM-powered self-supervision. Adding more unlabeled text (e.g., from PubMed) may further improve the performance.}
\begin{tabular}{llll}
\textbf{Task}  & \textbf{Method}    & \textbf{Model} & \textbf{Test F1} \\
\midrule
GAD            & LLM  & GPT-3.5 (few-shot)        & 49.25       \\
                & Distillation & PubMedBERT     & 56.42       \\
\hline
PHI(i2b2 2014) & LLM  & GPT-3.5 (few-shot)        & 64.20      \\
                & Distillation & PubMedBERT     & 73.89       \\
\hline
MedNLI         & LLM  & GPT-3.5 (few-shot)        & 82.21       \\
                & Distillation & PubMedBERT     & 80.24       \\
\bottomrule
\end{tabular}
  \label{tab:othertasks} 
\end{table}

As \Cref{tab:othertasks} shows, LLM distillation attains similar gains for GAD and PHI, which are both information extraction tasks not unlike ADE extraction. For MedNLI, however, GPT-3.5 slightly outperforms its distilled student model. This is not surprising, as MedNLI is a textual-entailment task, which is particularly suited for generative models like GPT. Moreover, for simplicity, we only use the unlabeled text from the training data (with labels removed) for distillation in these experiments. Better distilled models may be attained if we apply LLM self-supervision to a larger unlabeled dataset, as in ADE extraction.

\section{Discussion} 


In this study, we investigated the potential of using \acp{LLM} for scaling biomedical knowledge curation. We found that \acp{LLM}, such as GPT-4, already possess a reasonable capability in structuring biomedical text and substantial gains can be attained by distilling LLMs into task-specific student models through self-supervised learning. This approach provides additional advantages, such as efficiency, and white-box model access.

We conducted a case study on adverse drug event (ADE) extraction, a key health area in its own right. Our GPT-3.5 distilled PubMedBERT model achieved comparable accuracy to supervised state-of-the-art methods without using any labeled data. Despite being over 1,000 times smaller, the distilled model outperformed its teacher GPT-3.5 by over six absolute points in F1 and GPT-4 by over five absolute points.

Ablation studies on distillation model choice (e.g., PubMedBERT vs. BioGPT) and ADE extraction architecture shed light on best practices for biomedical knowledge extraction. Similar gains were attained by distillation for other standard biomedical knowledge extraction tasks, such as gene-disease associations and protected health information, further illustrating the promise of this approach.

These findings suggest that LLM distillation and domain-specific models, like PubMedBERT, can significantly contribute to the advancement of machine learning in healthcare. By harnessing the knowledge and capabilities of large language models, we can develop more efficient, cost-effective, and powerful solutions for various healthcare applications.


\paragraph{Limitations}

Despite the promising results, our study has several limitations:

Firstly, at the time of this work, the GPT-4 model has just been released. Due to time constraints, we did not conduct the distillation process using GPT-4 as the teacher model. In our few-shot setting, GPT-4 exhibited marginally better performance compared to GPT-3.5. Although we suspect that GPT-4 might be a better teacher, the expected gains are likely to be marginal.

Secondly, during the evaluation process, we assumed the presence of gold drug entities. This assumption is not held by several prior works that we compared our approach against. This difference in methodology might lead to a slight advantage in our setting, as our method relies on accurate drug entity identification to perform effectively.

Lastly, for knowledge distillation on other clinical tasks, we used the training corpus as input for the teacher model. However, given the relatively small size of these corpora, we have not been able to fully explore the true potential of distillation on these tasks. The limited data might restrict the effectiveness of the distillation process, and we acknowledge that there might be room for improvement with more extensive data and experimentation.

In summary, the limitations of our study include the use of GPT-3.5 instead of GPT-4 as the teacher model, the assumption of gold drug entities during evaluation, and the unexplored potential of distillation on other clinical tasks due to small training corpora. Future work could address these limitations by incorporating the latest language models, refining the evaluation process, and exploring the impact of larger training sets on knowledge distillation performance. 

\paragraph{Future Work}

To address the limitations and further enhance the performance of ADE extraction and other clinical tasks, several avenues for future research can be explored:

\begin{itemize}
\item \textit{Incorporating additional domain-specific knowledge sources}: Leveraging external domain-specific knowledge, such as ontologies and databases, could help improve model performance and address the issue of inconsistent annotations in the ADE dataset.
\item \textit{Expanding training corpus for other clinical tasks}: Increasing the training corpus for other clinical tasks using LLMs on unlabeled data could lead to improved performance in those tasks.
\item \textit{Evaluating on a broader range of clinical tasks and datasets}: Exploring the application of our proposed method on additional clinical tasks and datasets can provide further insights into the generalizability and adaptability of our approach in various healthcare contexts.
\item \textit{Investigating the use of GPT-4 in knowledge distillation}: Evaluating the potential benefits of incorporating GPT-4 in the knowledge distillation process could lead to further improvements in model performance across different clinical tasks.
\end{itemize}

\bibliography{ref_aiden}
\appendix
\section{Annotation Inconsistencies}
\label{sec:annotation-inconsistencies}


In this appendix section, we address the presence of annotation inconsistencies in the ADE corpus. Table \ref{tab:inconsistent-annotation} showcases examples of these inconsistencies, particularly in ambiguous boundaries, which can potentially impact the performance of machine learning models trained on this dataset. Researchers and practitioners should be cognizant of these inconsistencies when working with the ADE corpus to develop or assess their models.

\begin{table}[!ht]
\centering
\caption{Examples demonstrating inconsistencies in annotation criteria within the ADE corpus. ADE mention annotations are underlined, while discrepancies in the inclusion of similar words are shown in bold.}
\label{tab:inconsistent-annotation}
\begin{tabular}{p{\linewidth}}
\toprule
Examples \\
\midrule
\begin{minipage}[t]{\linewidth}
\begin{itemize}
\item CONCLUSIONS: Peripheral administration of low-dose vasopressin for septic shock should be discouraged because of the risk of \underline{ischemic skin \textbf{complications}}.
\item Warfarin-associated \underline{bleeding} \textbf{complication} saved life
\end{itemize}
\end{minipage} \\
\midrule
\begin{minipage}[t]{\linewidth}
\begin{itemize}
\item \underline{Acute pulmonary \textbf{reactions}} to nitrofurantoin are an uncommon side effect of therapy and can cause minor or life-threatening pulmonary dysfunction.
\item Several \underline{hypersensitivity} \textbf{reactions} to cloxacillin have been reported
\end{itemize}
\end{minipage} \\
\midrule
\begin{minipage}[t]{\linewidth}
\begin{itemize}
\item We stress the potential of benzarone to cause hepatotoxicity, which usually resembles \underline{\textbf{severe} chronic active hepatitis}.
\item Epoprostenol may be associated rarely with \textbf{severe} \underline{erythroderma}.
\end{itemize}
\end{minipage} \\
\midrule
\begin{minipage}[t]{\linewidth}
\begin{itemize}
\item In one patient the vasculitis resolved after termination of the ciprofloxacin therapy; in the other patient the ciprofloxacin-induced hemorrhagic vasculitis was superimposed on a severe forefoot infection, leading to \underline{\textbf{progressive} gangrene} and a below-knee amputation.
\item The potential for \textbf{progressive} \underline{brain injury} and subsequent disability related to intraventricular IL-2 therapy is discussed.
\end{itemize}
\end{minipage} \\
\midrule
\begin{minipage}[t]{\linewidth}
\begin{itemize}
\item \underline{\textbf{Lethal} anuria} complicating high dose ifosfamide chemotherapy in a breast cancer patient with an impaired renal function.
\item Late \textbf{lethal} \underline{hepatitis B virus reactivation} after rituximab treatment of low-grade cutaneous B-cell lymphoma.
\end{itemize}
\end{minipage} \\
\bottomrule
\end{tabular}
\end{table}

\end{document}